# Fast Trajectory Simplification Algorithm for Natural User Interfaces in Robot Programming by Demonstration

Daniel L. Marino, Milos Manic
Department of Computer Science
Virginia Commonwealth University
Richmond VA, USA
marinodl@vcu.edu, misko@ieee.org

*Abstract*— Trajectory simplification is a problem encountered in areas like Robot programming by demonstration, CAD/CAM, computer vision, and in GPS-based applications like traffic analysis. This problem entails reduction of the points in a given trajectory while keeping the relevant points which preserve important information. The benefits include storage reduction, computational expense, while making data more manageable.

Common techniques formulate a minimization problem to be solved, where the solution is found iteratively under some error metric, which causes the algorithms to work in super-linear time. We present an algorithm called *FastSTray*, which selects the relevant points in the trajectory in linear time by following an "open loop" heuristic approach. While most current trajectory simplification algorithms are tailored for GPS trajectories, our approach focuses on smooth trajectories for robot programming by demonstration recorded using motion capture systems.

Two variations of the algorithm are presented: 1) aims to preserve shape and temporal information; 2) preserves only shape information. Using the points in the simplified trajectory we use cubic splines to interpolate between these points and recreate the original trajectory. The presented algorithm was tested on trajectories recorded from a hand-tracking system. It was able to eliminate about 90% of the points in the original trajectories while maintaining errors between 0.78-2cm which corresponds to 1%-2.4% relative error with respect to the bounding box of the trajectories.

## I. Introduction

Nowadays we can record trajectory data from a wide range of sources like motion capture systems, touch screens, GPS, IMUs, cameras, among others. Processing this kind of data is a key point to different applications, which one among them is the construction of better natural user interfaces that are based on user input like gestures or demonstrations.

Trajectory data in general is recorded as a sequence of (position, time stamp)-pairs; In general, highly dense trajectories are obtained when tracking the movement of a person or an object. This is because of the high sampling ratio. Therefore, it is usual to simplify the raw data by reducing the number of points, a process called trajectory simplification [1], which allows to reduce the storage and further processing cost of the trajectories and makes the data more manageable. Hence, the problem of trajectory simplification is to get a compact, simple representation of the trajectory that preserves as close as possible the original trajectory.

The standard way of solving the problem is by the user providing to the algorithm a maximum error $\varepsilon$, then the algorithm sketches a simplified trajectory and measures the error between the original points and the simplified trajectory, if the error is too high the simplified trajectory is changed, in general by increasing its complexity adding a new point. One of the most well-known algorithms for trajectory simplification is the Ramer–Douglas–Peucker algorithm, the algorithm run time is $O(n \log(n))$ and $O(n^2)$ in the worst case, a modification of this algorithm is given in [2].

The approach mentioned in the previous paragraph is the general design paradigm for algorithms that solve the trajectory simplification problem. We refer to these algorithms as "closed loop" because the choice of the points that will be kept is done based on testing the sketch of a simplified trajectory over an error metric, and modifying this trajectory according to the feedback given by the error metric. This closed loop nature is the main reason for the algorithms to have super-linear running time.

In this paper we propose an algorithm called *FastSTray* (Fast simplification trajectory), to perform trajectory simplification. The algorithm is able to select the relevant points in a trajectory in linear time $O(n)$, where $n$ is the number of points on the original trajectory. The algorithm was developed specifically to process trajectories acquired from the tracking of human demonstrations using motion capture systems [3]. These demonstrations are used to specify trajectories for robot manipulators that execute the demonstrated trajectories, a process called programming by demonstration [4] [5].

*FastSTray* calculates for each point on the original trajectory a coefficient that quantify its importance on the definition of the trajectory, we call this coefficient "information" coefficient; the calculation of the coefficient is done in linear time. Once the coefficient is calculated for all points, *FastSTray* selects which points should be kept by applying a non-maxima suppression to eliminate points with low coefficients. Using only the remaining points, the algorithm calculates the parameters of a cubic spline, which gives us an approximation of the original trajectory. The fact that the points are being selected without constantly evaluating the error of the simplified trajectory with respect the original points is which gives our algorithm its "open loop" attribute and allows it to run in linear time.

Using splines to interpolate between points in the simplified trajectory, gives us a smooth trajectory that is suitable for robotics, and it also provides a better model for the type of data that we are working on (tracking of human actions) that allows the simplified trajectory to fit the original trajectory with low error and high reduction ratios.

Another reason of the use of splines is that they are easy to interpret and modify by humans. As mentioned before, we are using the trajectories recorded of a human performing an action to serve as a demonstration for a robot of the trajectory it should follow to execute a task; it is desired that this trajectory would be in a format that it would be easy for a human to modify in case that a modification of the trajectory or fine tuning is needed; splines are often used to program robot trajectories, and they are intuitive and easy to modify on OLP software [6].

We conducted tests on data gathered from a hand-tracking system [3] and on GPS trajectory data. The results show a reduction ratio about 90%, with relative error between 0.5%-2.5%.

This paper is organized as follows: in section II a review on the related work on trajectory simplification and similar areas is presented; section III explains the *FastSTray* algorithm for trajectory simplification using a linear correlation coefficient that aims to preserve temporal and shape information; in section IV a coefficient that aims to preserve only shape information is presented; in section V the results on the test trajectories are shown; section VI concludes the paper.

## II. RELATED WORK

The development of trajectory simplification algorithms has been an important topic specially for the processing of GPS trajectory data. In [7], long et al. classify the existing approaches in *position-preserving trajectory simplification* (PPTS) [8], and *direction-preserving trajectory simplification* (DPTS) [1]. PPTS algorithms ensure that the error between the positions on the simplified trajectory and the original trajectory is less than a certain given parameter; whereas DPTS tries to preserve the direction information on the data. In PPTS and DPTS algorithms the goal is to end up with the smallest possible simplified trajectory while the error is held below a certain threshold, in [7] this problem is called the *min-size* problem, because the goal is to minimize the size of the trajectory. An issue found with this approach is that the user has to specify the error threshold which sometimes is hard to tune, therefore [7] defines the *min-Error* problem which takes as user parameter a *storage budget* which is more intuitive to specify than the error threshold, but the drawback is the computational cost which according to [7] it is $O(n^2 \log(n))$ for the exact algorithm, and $O(n \log^2(n))$ for the approximate algorithm.

In [9] an online trajectory sampling method for portable devices is presented to reduce the trajectory data resulting from streaming location data from portable devices to a location-based services (LBS) server. The approach used is similar to the Ramer–Douglas–Peucker algorithm, but it assumes that the entire trajectory is not known a priori, hence the simplification of the trajectory is performed online while the data is being acquired.

In [10] a trajectory simplification algorithm specially tailored for GPS trajectories is presented, this algorithm was designed to preserve the shape and semantic meaning of the GPS trajectory; it also shares some of the ideas that we use in our work like using the deviation in heading direction to weight the importance of points and make the decision of which points should be kept according to this weight.

Another interesting area that is related is the problem of fitting spline curves to unorganized data points [11]. The standard approach is to formulate the problem as a nonlinear constrained minimization problem. This problem is more difficult that the trajectory simplification problem because it assumes that the data is unordered. Calculating the error between the current estimate of the spline and the original data points is one of the critical steps for fitting the spline. Finding the exact value for the error is prohibitively expensive, therefore an approximation has to be used. There are three mainly existing methods for curve fitting that propose different ways of approximating the error: the first method is called Point Distance Minimization (PDM) [12]; the second method is called tangent distance minimization (TDM) [13] and the third one is the Squared Distance Minimization (SDM) [14]. The PDM method is the simplest one, but the SDM exhibits better performance in terms of stability and convergence rate than PDM and TDM; furthermore, in [14] the SDM method is shown to be a quasi-Newton method which uses a positive definite approximant of the Hessian of the objective function that defines the fitting problem. In [11] a method that uses L-BFGS optimization is used that is faster than the other methods.

## III. DESCRIPTION OF THE PRESENTED FASTSTRAY ALGORITHM

This section describes the algorithm using a linear-correlation-based coefficient; later on, in section IV, the direction-based coefficient is explained. The presented algorithm takes as input a trajectory $T$ and returns trajectory $\hat{T}$. The output trajectory contains fewer points and can be used to recreate the trajectory $T$ using splines. Figure 1 shows the main steps of the algorithm, while detailed procedure is described in Algorithm 1. $\alpha, \beta,$ and $\gamma$ are parameters given by the user who also has to specify which coefficient should be used.

The input to the algorithm is a raw trajectory data $T$ which is a structure composed by a sequence of N ordered points $T.P = \{p(1), p(2), \dots, p(N)\}$, and the corresponding time stamps for each point $T.S = \{t(1), t(2), \dots, t(N)\}$, where the positions $p(i) \in \mathbb{R}^3$ and the time stamp $t(i) \in \mathbb{R}$.

The output is a trajectory $\hat{T}$, which is an approximation of the input trajectory $T$, with $M$ points, where $M < N$ and $\hat{T}.\hat{P} = \{p(1), p(2), \dots, p(M)\}$, $T.\hat{S} = \{t(1), t(2), \dots, t(M)\}$. The points on $\hat{T}$ are used to get a spline as specified in [15].

The first phase of the algorithm applies a filter to smooth the trajectory and reduce the noise. This step is completely optional and is used in cases where the trajectory provided is noisy. The filter used in this paper is a simple moving average filter with a window size $2\alpha$.

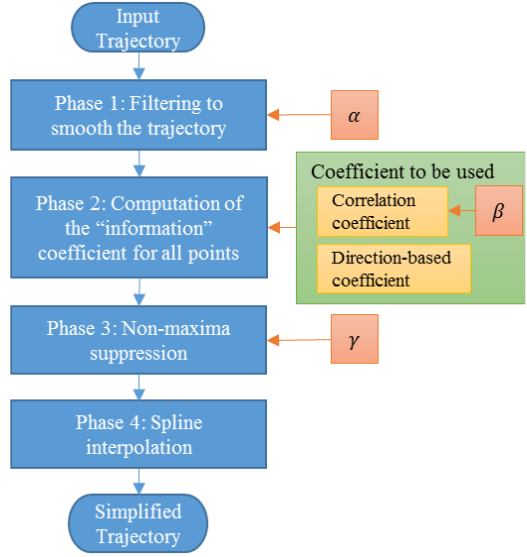

Figure 1: Steps performed by the presented algorithm

The second phase performs a measure of the amount of "information" that a point is providing to define the trajectory; we quantify this information by giving each point in the trajectory a coefficient $\xi \geq 0$, where a point that is providing significant information to reconstruct the trajectory should have a big coefficient. We introduce a linear-correlation-based coefficient *(details explained in Algorithm 1)*.

$$r_{at}(\{a\},\{t\}) = \frac{\sum_i (a(i) - \bar{a})(t(i) - \bar{t})}{\sqrt{\sum_i (a(i) - \bar{a})^2} \sqrt{\sum_i (t(i) - \bar{t})^2}} \quad (1)$$

$$\xi(\mathbf{P},t) = \frac{1}{\left(r_{at}(\mathbf{P}.x,t)\right)^2} + \frac{1}{\left(r_{at}(\mathbf{P}.y,t)\right)^2} + \frac{1}{\left(r_{at}(\mathbf{P}.z,t)\right)^2} \quad (2)$$

where $\mathbf{P}.x$ is a list of the x coordinate of the list of points $\mathbf{P}$, similarly $\mathbf{P}.y$ and $\mathbf{P}.z$ represent y and z coordinate respectively.

Figure 2 illustrates the idea behind the design of this coefficient. We would like eliminate points with low curvature because they do not provide significant information for the definition of the trajectory, like the one highlighted in Figure 2.a), and to preserve points like the one highlighted in Figure 2.b). To helps us to identify these points, we are using the correlation coefficient between the $2\beta$ neighbors of the point whose coefficient is being evaluated, as shown in Figure 2 (where the number of neighbors in this figure is $2\beta = 4$), points with low linear correlation are those that we would like to preserve, so we assign a high coefficient to these points by taking the multiplicative inverse of the squared linear-correlation of position with respect to time.

Phase 3 eliminates points with low $\xi$ coefficient by performing a non-maxima suppression, preserving only points that are key to reproduce the original trajectory. The non-maxima suppression works as follows: a point is only preserved if its "information" coefficient corresponds to the maximum coefficient found on a window of size $2\gamma$ centered on the point whose coefficient is being evaluated, i.e. a point is preserved if its coefficient has the maximum value on its $\gamma$-neighborhood.

Phase 4 calculates the parameters of a cubic spline using only the reduced set of points in the simplified trajectory $\hat{T}$.

Step 1 on Algorithm 1 corresponds to Phase 1 on Figure 1. Step 2 corresponds to Phase 2. Step 3 to 9 corresponds to Phase 3. Finally step 10 performs the Spline interpolation. Steps 3 and 9 in Algorithm 1 ensure that the initial and final point on the original trajectory belong to the simplified trajectory.

With respect to the computational complexity of the algorithm, the following are the runtimes of each one of the steps on Algorithm 1:

- Step 1: $O(\alpha N)$
- Step 2: $O(\beta N)$
- Steps 4-9: $O(\gamma N)$

Because $\alpha, \beta$ and $\gamma$ are in general much more smaller than N, we conclude that the running time of the algorithm for getting the simplified trajectory is $O(N)$. This running time corresponds only to the simplification of the trajectory $T$, it does not take into account the cost of getting the spline $\Pi$.

The spline $\Pi$ is a cubic interpolation between the points on $\hat{T}$, in other words $\Pi$ is composed by $|\hat{T}| = M - 1$ cubic polynomials $\Pi_k(t)$. To get the parameters of the $M - 1$ polynomials, a system of linear equations $A\Psi = b$ must be solved, where $A$ is a known $M$x$M$ matrix, $b$ is a known vector and $\Psi$ is the unknown vector that represent the acceleration on each one of the timestamps, please refer to [15] for the details of how to get $A$ and $b$. The simplest way for solving $A\Psi = b$ requires the inversion of the matrix $A$, which takes $O(M^3)$; the reduction ratio that was achieved in the tests that we performed was about 90%, therefore $M \approx 0.1N$, which means that the running time for the inversion of $A$ is about $0.001N^3$. This is a significant reduction in the running time even when $N$ is large; furthermore, for cubic interpolation, the matrix A has a

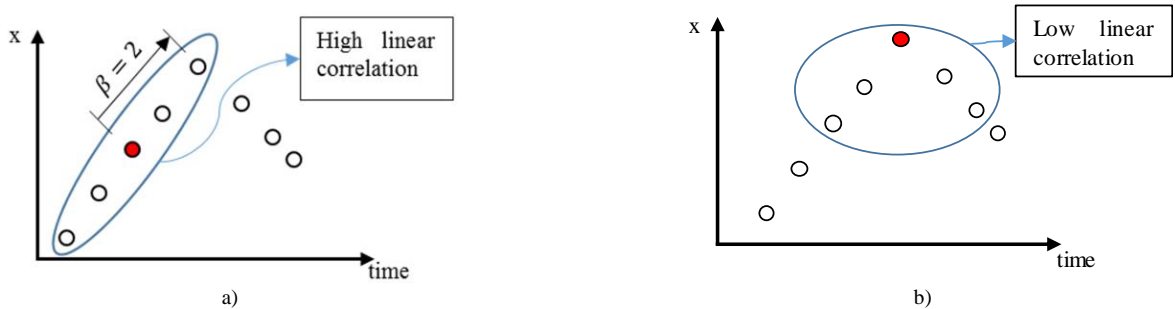

Figure 2: examples that demonstrate the central idea for the presented "information" coefficient based on linear correlation

tridiagonal band structure, therefore the linear system of equations can be solved in O(M) using the Tridiagonal matrix algorithm (Thomas algorithm). There are also efficient algorithms like [16] for calculating the inverse of A.

---

**Algorithm 1: FastSTray (T, α, β, γ)**

**Input:** Trajectory $T$, composed of a list of points $P$ and a list of time stamps $S$

**Output:** Simplified trajectory $\hat{T}$, composed of the list of points $\hat{P}$ and a list of time stamps $\hat{S}$.
Spline $\Pi(t)$ that interpolates points in $\hat{T}$

**Parameters:** $\alpha$: Size of the moving average filter
$\beta$: Size of the neighborhood to measure the correlation coefficient
$\gamma$: Size of the neighborhood to perform the non-maximum suppression

1. Calculate the trajectory $T_1$(composed by a list of points $P_1$ and time stamps $S_1$) using moving average filter:

   $P_1[i] \leftarrow \dfrac{\sum_j P[j]}{|J|}$,
   $S_1[i] \leftarrow S[i]$
   Where $J = \{j \in \mathbb{N} | \max(0, i - \alpha) \leq j \leq \min(i + \alpha, N)\}$

2. For each point $P_1[i]$ get the coefficient $Coef(i)$ that corresponds to the measure of the linear correlation of the $2\beta$ neighbors of $P_1[i]$ with respect to time, according to the correlation measure ($\xi$) defined in Eq. (2)

   $Coef(i) \leftarrow \xi(P_v(i), t_v(i))$
   Where:
   $P_v(i) = \{P_1[j] | i - \beta \leq j \leq i + \beta\}$
   $t_v(i) = \{S_1[j] | i - \beta \leq j \leq i + \beta\}$

3. $\hat{P} = P_1[1]$ ; $\hat{S} = S_1[1]$, i.e. add to the simplified trajectory the initial point of the filtered trajectory

4. **For** each point $P_1(i)$ :
5.   $M_p \leftarrow \max(\{Coef(j) | i - \gamma \leq j \leq i + \gamma\})$
6.   **If** $Coef(i) = M_p$ **Then**
7.     $\hat{P} \leftarrow \hat{P} \cup P_1[i]$, i.e. add the point to the simplified trajectory
8.     $\hat{S} \leftarrow \hat{S} \cup S_1[i]$
9. $\hat{P} \leftarrow \hat{P} \cup P_1[N]$ ; $\hat{S} \leftarrow \hat{S} \cup S_1[N]$, i.e. add to the simplified trajectory the final point of the filtered trajectory
10. Return $\hat{T}$ as a structure composed by $\hat{T}.\hat{P}$ and $\hat{T}.\hat{S}$, and use the ordered points in the trajectory $\hat{T}$ to get the spline $\Pi$.

---

## IV. PRESERVING ONLY SHAPE INFORMATION

The coefficient defined in Eq. (2) aims to preserve position information and temporal information, this might be desired in certain applications like in robot programing by demonstration, where temporal information provides important information on how the task has to be performed.

In applications like GPS trajectory simplification, temporal information is not relevant, only shape information is required, so we introduce a coefficient that aims to preserve only shape information. We call this coefficient *direction-based* coefficient.

Figure 3 illustrates the idea behind the *direction-based* coefficient. The coefficient makes use of the cosine similarity between the vectors v1 and v2, defined by:

$$\frac{v_1(i)^T v_2(i)}{\|v_1(i)\|_2 \|v_2(i)\|_2} \quad (3)$$

The cosine similarity takes values between [-1,1], where $v_1$ and $v_2$ have similarity of 1 if they have the same orientation; a similarity of zero corresponds to perpendicular vectors; and a similarity of -1 corresponds to a pair of vectors that are diametrically opposed. Based on this behavior of the similarity measure, we define the "information" coefficient as:

$$Coef(i) = \frac{1}{1 + \dfrac{v_1(i)^T v_2(i)}{\|v_1(i)\| \|v_2(i)\|}} \quad (4)$$

where:

$$v_1(i) = P_1[i] - P_1[i-1]$$
$$v_2(i) = P_1[i+1] - P_1[i] \quad (5)$$

The idea is that the inner product between the vectors shown in Figure 3.b) will be higher than the inner product of the vectors shown in Figure 3.a). which allow us to choose the point in Figure 3.a) over the point in Figure 3.b) to be kept. The only modification that must be done on Algorithm 1 is to change step two to use the coefficient defined in Eq. (4) instead of using the coefficient defined in Eq. (2)

It is important to highlight the fact that in Figure 1 the axes are showing position vs time, whereas in Figure 3 the axes show position x vs position y. In section III the coefficient was defined in base of the linear correlation between the individual coordinates with respect to time, this is the main reason of why this coefficient preserves temporal information, while the coefficient presented in this section measures the difference in the direction between two adjacent points.

## V. EXPERIMENTAL RESULTS

For evaluating the performance of the presented algorithm, the following error metrics is being used:

$$error = \frac{1}{N} \sum_{i=1}^{N} \|P[i] - \Pi(T.t(i))\| \quad (6)$$

where $P[i]$ represents the points in the original trajectory; and $\Pi$ represents the spline obtained using the points in the simplified trajectory $\hat{T}$, where $\Pi(t)$ corresponds to position on 3D space interpolated via spline at time $t$.

This metric can be considered a *synchronous distance function* [7], because the points in the original trajectory $T$ are mapped to the spline (defined with the points in $\hat{T}$) using the same time stamp in $T.t$.

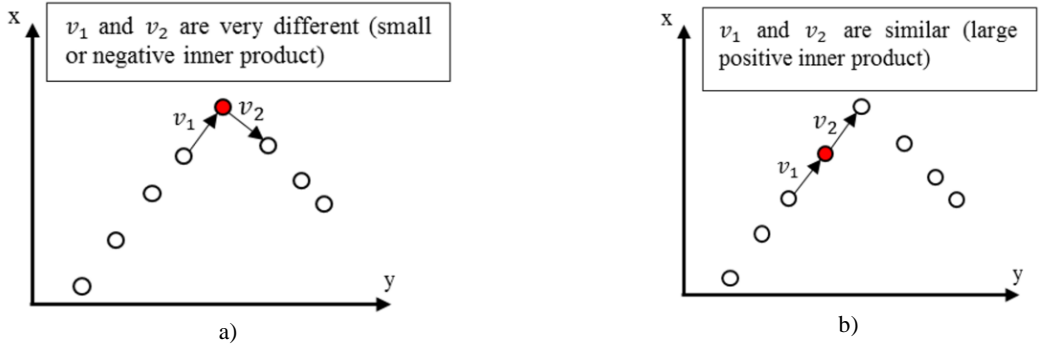

Figure 3: Visual example that demonstrates the idea behind of the presented direction-based coefficient that only preserve shape information

The performance is also being evaluated by the percentage of reduction achieved, which we define as:

$|T|$ = Number of points in the original trajectory
$|\hat{T}|$ = Number of points in the simplified trajectory

$$Reduction(\%) = 100\left(1 - \frac{|\hat{T}|}{|T|}\right) \quad (7)$$

We also define a relative error as the ratio between the error defined in Eq. (6) and de maximum distance between two points (in 3D space) in the original trajectory:

$$relative\ error = 100 \frac{error}{\max_{i,j}(\|P[i] - P[j]\|)} \quad (8)$$

Figure 4 presents the results of the presented algorithm using the linear-correlation-based coefficient. The trajectories that we used for testing were recorded using a hand-tracking system [3]. As can be seen, this kind of trajectories are characterized by a high sampling ratio and represent the movement of an object over a smooth trajectory over time (there are no discontinuities in the data). We can see that the algorithm is producing simplified trajectories with reduction rates about 90%, and errors between 0.78cm to 2cm, which gives relative errors between 1% to 2.4%.

Figure 5 presents the results of the *FastSTray* algorithm using the *direction-based* coefficient on a trajectory that tries to resemble a square. As can be seen, the higher reduction rate, the higher the error; we can also see that the coefficient $\gamma$ is used to control and tune the amount of reduction rate. Here, high $\gamma$ values results in high reduction rates and higher mapping errors.

It is important to highlight the fact that for the tests with the direction-based coefficient we are still using the *synchronous distance function* error. Due to the fact that the direction-based coefficient is not preserving temporal function, the error metric reports higher errors compared to the results got with the correlation-based coefficient because of the nature of the error metric.

To test the algorithm on a different kind of trajectory data we used the GeoLife GPS Trajectories dataset [17] [18]. Table 1 illustrates mapping error and reduction percentages for $\alpha = 5$ and varying $\gamma$ values. As can be seen, the error increases as $\gamma$ increases. These results were obtained by applying the *FastSTray* algorithm using the *direction-based* coefficient on a GPS trajectory with 3189 points.

TABLE I
PERFORMANCE OF THE PRESENTED ALGORITHM ON A GPS TRAJECTORY WITH $|T| = 3189$ AND USING $\alpha = 2$

| $\gamma$ | $|\hat{T}|$ | Reduction percentage | Error [m] | Relative Error [%] |
|---|---|---|---|---|
| 1 | 420 | 86.83 | 10.6 | 0.5 |
| 2 | 274 | 91.4 | 15.6 | 0.74 |
| 3 | 197 | 93.82 | 24.14 | 1.14 |
| 4 | 153 | 95.2 | 29.15 | 1.38 |
| 5 | 122 | 96.17 | 32.78 | 1.55 |
| 6 | 105 | 96.72 | 38.35 | 1.82 |

The GPS trajectory with 3189 points also served as evidence of the fast performance of the algorithm, which took 0.06 seconds to the get the simplified trajectory using the direction-based coefficient, and 0.33 seconds using the linear-correlation coefficient, in a non-optimized implementation of the algorithm in Matlab, on a computer with an AMD 4GHz processor.

When working with GPS trajectories, special care has to be taken with discontinuities in the trajectory data, the results that we are showing are not taking into account these discontinuities to process the trajectory which leads to poor performance in the regions where discontinuities are found.

Another important point found when working with GPS trajectories, is that with very long trajectories, using the same values for $\alpha, \beta$ and $\gamma$ for the entire trajectory, usually leads to poor performance in local regions. For example, when there is a segment with higher variance compared to the average behavior of the trajectory, the spline would have a poor fit in this local region. A simple approach to improve performance for these cases is to break the trajectory in segments and use different values of $\alpha, \beta$ and $\gamma$ for each segment.

The fact that we are using splines instead of linear interpolation allows us to have higher reduction ratios (compared to linear interpolation) while the reconstructed trajectory fits the original data points with low error. It is important to highlight that the fitting of the spline is not taking into account the points in the original trajectory $T$, the spline

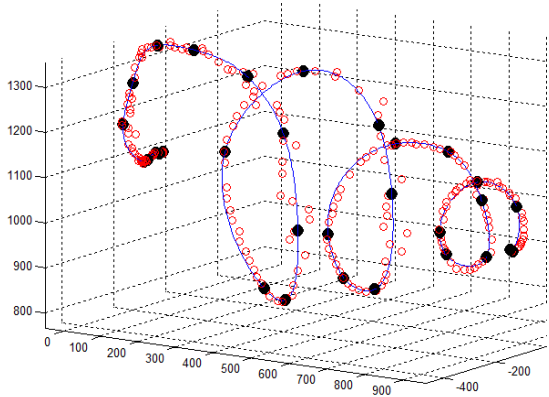
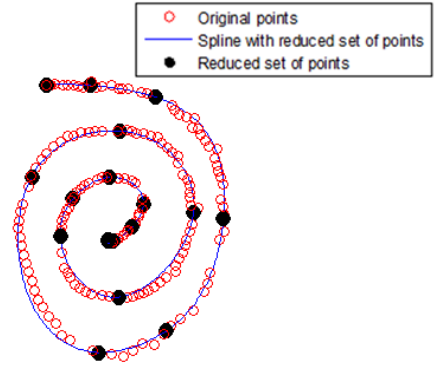

|T| = 334, |T̂| = 31, Reduction= 91%, error: 9.5mm, Relative error= 0.9%, $\alpha = 1, \beta = 2, \gamma = 2$

|T| = 270, |T̂| = 18, Reduction= 93.3%, Error: 13.4mm, Relative error= 1.66%, $\alpha = 1, \beta = 2, \gamma = 3$

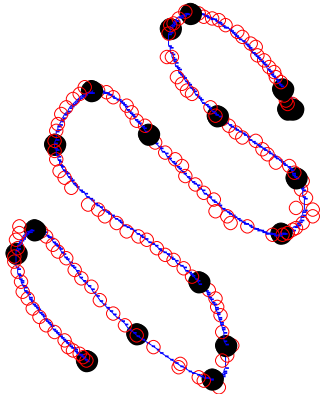
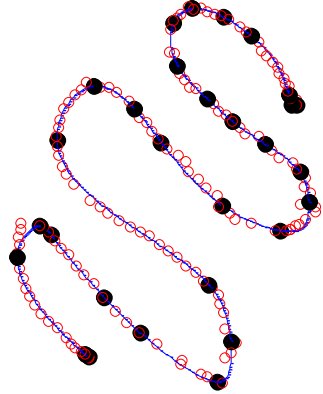
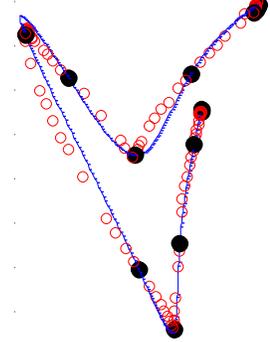

|T| = 219, |T̂| = 18, Reduction= 91.78%, Error: 15.9mm, Relative error= 2.35%, $\alpha = 2, \beta = 3, \gamma = 2$

|T| = 219, |T̂| = 30, Reduction= 86.3%, Error= 7.8mm, Relative error= 1.15% $\alpha = 1, \beta = 2, \gamma = 1$

|T| = 181, |T̂| = 16, Reduction= 91.16%, error= 15mm, Relative error= 1.99%, $\alpha = 1, \beta = 2, \gamma = 2$

Figure 4: Results of the trajectory simplification algorithm using the correlation coefficient for trajectories recorded from a human hand tracking system

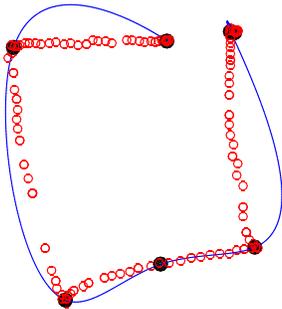
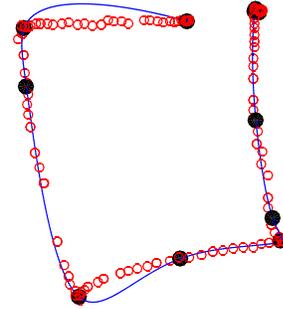

|T| = 207, |T̂| = 7, Reduction= 96.6%, error: 57.84mm, $\alpha = 5, \gamma = 5$

|T| = 207, |T̂| = 11, Reduction: 94.68%, error: 40.85mm, $\alpha = 5, \gamma = 3$

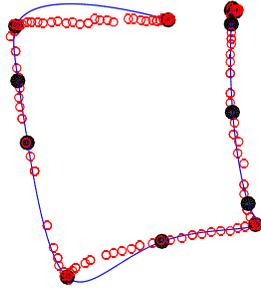
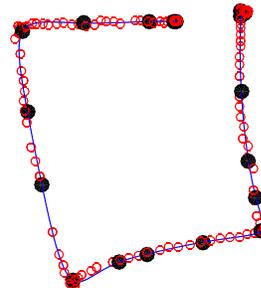

|T| = 207, |T̂| =16, Reduction: 92.27%, error: 37.18mm, $\alpha = 5, \gamma = 2$

|T| = 207, |T̂| = 19, Reduction: 90.82%, error: 54.26mm, $\alpha = 8, \gamma = 1$

Figure 5: Result of the trajectory simplification algorithm using the direction-based coefficient on trajectory data recorded using a hand-tracking system.

interpolation is only taking into account the points in the simplified trajectory $\hat{T}$, nonetheless the results show a low error in the mapping of the original trajectory to the spline, which means that the coefficients designed in this paper are a good heuristic for the "open loop" approach that we took to solve the trajectory simplification problem. If the error would like to be further reduced, one approach could be to use the spline Π as an initialization for a "closed loop" algorithm like [11].

Although the algorithm requires three parameters to be specified, the tests performed showed that their values usually fall in the range [1,10], mainly because values higher than 10 yield to high reduction ratios and high errors.

The choice of the parameters $\alpha, \beta$ and $\gamma$ is intuitive and its tuning is easy because their value correspond to the size of neighborhoods for the corresponding operations, therefore the parameters do not depend on the scale of the data. On the other hand, the fast performance of the algorithm allows fine tuning of the parameters interactively.

Another fact that we observed while performing the tests is related to the role of the smoothing filter operation. We found on our tests that big values of $\alpha$ lead to bigger reduction ratios and sometimes to lower mapping errors. Therefore, even if the input trajectory is noise-free, this operation still can aid to a better detection of significant points that should be preserved.

## VI. CONCLUSION

This paper presents the *FastSTray* algorithm for trajectory simplification on smooth trajectories. The main advantage of the algorithm is that is able to select the relevant points that allows to recreate the original trajectory using splines with low error and high reduction ratios in linear time, showing that the proposed coefficients are a good heuristic for the problem by producing simplified trajectories with reduction rates about 90%, and errors between 0.78cm to 2cm, which gives relative errors between 1% to 2.4%, for trajectories recorded from a hand-tracking system.

Another advantage of the algorithm is that the required parameters are intuitive and easy to tune, mainly because the parameters define the neighborhood used by each one of the phases of the algorithm, therefore they are invariant to the scale of the trajectories. On the other hand, because the algorithm runs in linear time, and the parameters usually take values on a small range, algorithms that automate the choose of these parameters can be proposed.

In the case that lower error rates need to be achieved, the simplified trajectory obtained with the presented algorithm can be used as an initialization for an algorithm that takes a standard "closed loop" minimization-optimization approach for fitting the spline on the points in the original trajectory.

In general, *FastSTray* works better for smooth trajectories with constant high sampling ratio. This is basically because the algorithm was specifically designed to work on smooth trajectories used in natural interfaces for robot programming by demonstration environments, nonetheless, we showed that the algorithm could also be used on GPS trajectories although further considerations should be taken on this kind of data, especially when there is high variance in the sampling ratio and/or spacing on neighbor points.

As future work, we will consider: 1) the orientation for the end-effector. 2) longer trajectories where the values for the coefficients $(\alpha, \beta, \gamma)$ will vary along the trajectory to improve the fitting of the spline in local regions.